\documentclass[10pt,twocolumn,letterpaper]{article}

\usepackage{iccv}              

\definecolor{iccvblue}{rgb}{0.21,0.49,0.74}
\usepackage[pagebackref,breaklinks,colorlinks,allcolors=iccvblue]{hyperref}

\def\paperID{7819} 
\def\confName{ICCV}
\def\confYear{2025}

\title{When Schrödinger Bridge Meets Real-World Image Dehazing with Unpaired Training}
\author{Yunwei Lan\textsuperscript{1, 2}, Zhigao Cui\textsuperscript{1*}, Xin Luo\textsuperscript{2}, Chang Liu\textsuperscript{2}, Nian Wang\textsuperscript{1},\\
Menglin Zhang\textsuperscript{2}, Yanzhao Su\textsuperscript{1}, Dong Liu\textsuperscript{2*}\\
\textsuperscript{1}Rocket Force University of Engineering, Xi’an 710025, China\\
\textsuperscript{2}MOE Key Laboratory of Brain-Inspired Intelligent Perception and Cognition, \\
University of Science and Technology of China, Hefei 230093, China\\
}

\begin{document}
\maketitle
\begin{abstract}

Recent advancements in unpaired dehazing, particularly those using GANs, show promising performance in processing real-world hazy images. However, these methods tend to face limitations due to the generator's limited transport mapping capability, which hinders the full exploitation of their effectiveness in unpaired training paradigms. To address these challenges, we propose DehazeSB, a novel unpaired dehazing framework based on the Schrödinger Bridge. By leveraging optimal transport (OT) theory, DehazeSB directly bridges the distributions between hazy and clear images. This enables optimal transport mappings from hazy to clear images in fewer steps, thereby generating high-quality results. To ensure the consistency of structural information and details in the restored images, we introduce detail-preserving regularization, which enforces pixel-level alignment between hazy inputs and dehazed outputs. Furthermore, we propose a novel prompt learning to leverage pre-trained CLIP models in distinguishing hazy images and clear ones, by learning a haze-aware vision-language alignment. Extensive experiments on multiple real-world datasets demonstrate our method's superiority. Code: https://github.com/ywxjm/DehazeSB.
\footnote{We acknowledge the support of GPU cluster built by MCC Lab of Information Science and Technology Institution, USTC. Corresponding authors: Zhigao Cui and Dong Liu. (ywlan, xinluo, lc980413, zhangmenglin)@mail.ustc.edu.cn, cuizg10@126.com, nianwang04@outlook.com, syzlhh@163.com, dongeliu@ustc.edu.cn}
\end{abstract}   
\section{Introduction}
\label{sec:intro}

\begin{figure}[t]
\centering
\includegraphics[width=0.8\linewidth]{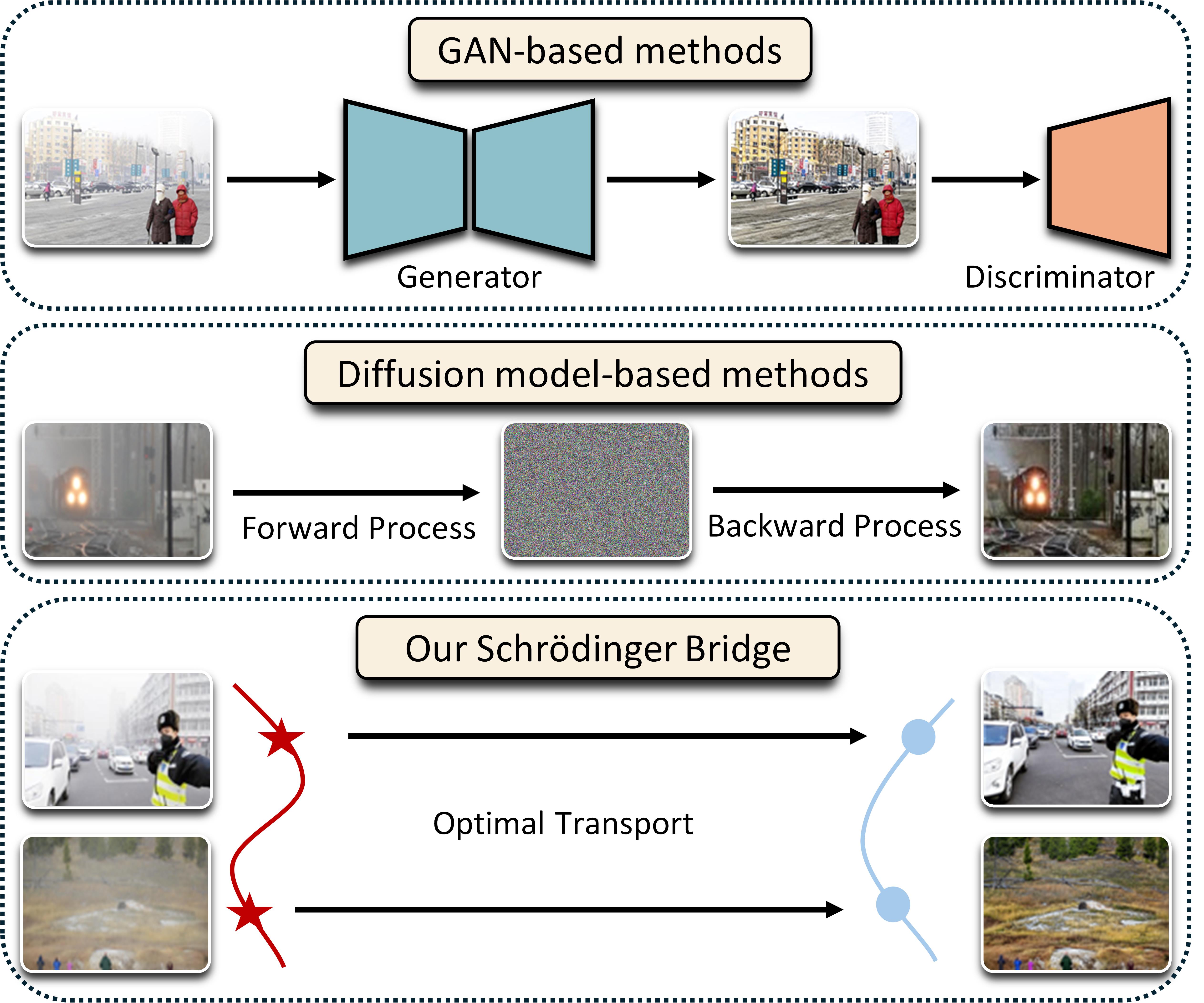}
\caption{\textbf{Existing GAN-based unpaired dehazing methods} struggle to handle images with complex structures due to the limited mapping capability of the generator. While \textbf{diffusion model-based methods} can achieve more intricate transport mappings, they necessitate transforming images into Gaussian noise and then regenerating them through a large number of sampling steps. \textbf{Our Schrödinger Bridge}  enables the direct learning of optimal transport mapping from hazy to clear image distributions in fewer time steps, achieving high efficiency and high-quality dehazing.}
\label{figure_1}
\end{figure}
Images captured under hazy conditions usually display reduced brightness and diminished contrast, significantly degrading visual quality and hindering further applications in visual tasks such as object detection \cite{farhadi2018yolov3}. Consequently, image dehazing, which aims to restore clear images from hazy ones, has received considerable attention over the past decade.
The degradation process of hazy images typically adheres to the Atmospheric Scattering Model (ASM) \cite{nayar1999vision}. 
\begin{equation}
\label{eq_1}
I(x)=J(x)t(x)+A(1-t(x)),
\end{equation}
where $I(x)$ represents the hazy image, and $J(x)$ is the corresponding clear one. $A$ and $t(x)$ are the atmospheric light and transmission map. 

Early studies \cite{he2010single,zhu2015fast,berman2016non} primarily rely on physical priors to estimate the parameters necessary for the ASM, subsequently obtaining clear images.  
Although these methods can effectively dehaze images, they are susceptible to over-dehazing since hand-crafted priors are often scenario-specific. Recently, the advent of deep learning has spurred a shift towards utilizing deep learning for dehazing \cite{cai2016dehazenet,dong2020multi,zheng2023curricular,qiu2023mb,feng2024kanet, li2024ustc}.
These methods typically utilize synthetic paired data for training since real-world hazy and clear image pairs are normally tough to collect.
Although improved performance is observed, such paired training paradigm often fails to generalize to real-world scenarios, due to the ill-presenting performance of trained networks that lack real-world information from hazy images.
To tackle the bottleneck of paired training, recent studies \cite{zhao2021refinednet,yang2022self,li2021yoly,wang2024ucl} explore unpaired training for dehazing without the requirements of data pairs. 
This paradigm enables existing model architectures to be directly trained on real-world data, making it more suitable for real-world image dehazing.

As shown in Figure \ref{figure_1}, existing unpaired dehazing methods \cite{zhao2021refinednet, engin2018cycle, wang2024ucl, yang2022self} predominantly leverage Generative Adversarial Networks (GANs) \cite{goodfellow2020gan} to generate dehazed images. 
Typically, these methods first produce preliminary dehazed results using a generator, often based on an Encoder-Decoder. They then align the distribution of these results with that of unpaired clear images using a discriminator. However, these methods often struggle to handle images with complex and detailed structures due to limitations in the generator's ability to model transport mappings.
Recent advancements in diffusion models \cite{ho2020denoising, song2021denoising, rombach2022high} offer an effective alternative. Diffusion models can simulate more intricate transport mappings through an iterative process of adding and removing noise, generating diverse and high-quality dehazed results.
Although this capability is difficult for GANs to match, diffusion models must first transform the image into Gaussian noise and then generate the image through a large number of sampling steps. This process imposes limitations on their potential for unpaired dehazing and leads to inferior computational efficiency.

Another key challenge in unpaired dehazing is the way to ensure the structural consistency between hazy and dehazed images, so as to suppress misaligned information caused by unpaired clear images.
While CycleGAN \cite{zhu2017unpaired} serves as a classical solution and has been widely used in previous unpaired dehazing methods \cite{engin2018cycle, yang2022self, yang2024robust, lan2025exploiting}, it requires additional training of generators and discriminators, leading to additional computational cost. 
Additionally, we observe that a pre-trained CLIP model \cite{clip} can effectively distinguish between hazy and clear images. This insight motivates us to explore improving dehazing by maximizing the distance between hazy textual prompts and dehazed images. 
However, we find that directly providing textual descriptions is not the optimal solution. This may be due to the fact that the concentration and types of haze in real-world images 
are extremely complex, and simple textual descriptions fail to capture these characteristics.

To address the aforementioned challenges, we propose a novel framework for real-world image dehazing, namely DehazeSB, consisting of the following contributions:
\begin{itemize}
    \item We apply Schrödinger Bridge (SB) in our unpaired dehazing framework. Leveraging SB based on optimal transport (OT) theory, we establish direct bridges between hazy and clear image distributions. This enables us to find optimal transport mappings from hazy to clear images, thereby improving dehazing performance. Moreover, SB can generate high-quality images in just a few or even one step, significantly reducing computational overhead.
    \item To enforce a structure-consistent constraint while minimizing training overhead, we introduce detail-preserving regularization. This ensures content and texture consistency during the dehazing process.
    \item  We propose a novel prompt learning strategy to tailor the textual representations for hazy images. This learned prompt plays a more effective role in representing hazy images instead of direct textual descriptions.
    \item Extensive quantitative and qualitative experiments show that our method surpasses existing state-of-the-art dehazing methods, including paired or unpaired training.
\end{itemize}
\section{Related Work}
\label{sec:related}
\begin{figure*}[t]
\centering
\includegraphics[width=0.9\textwidth]{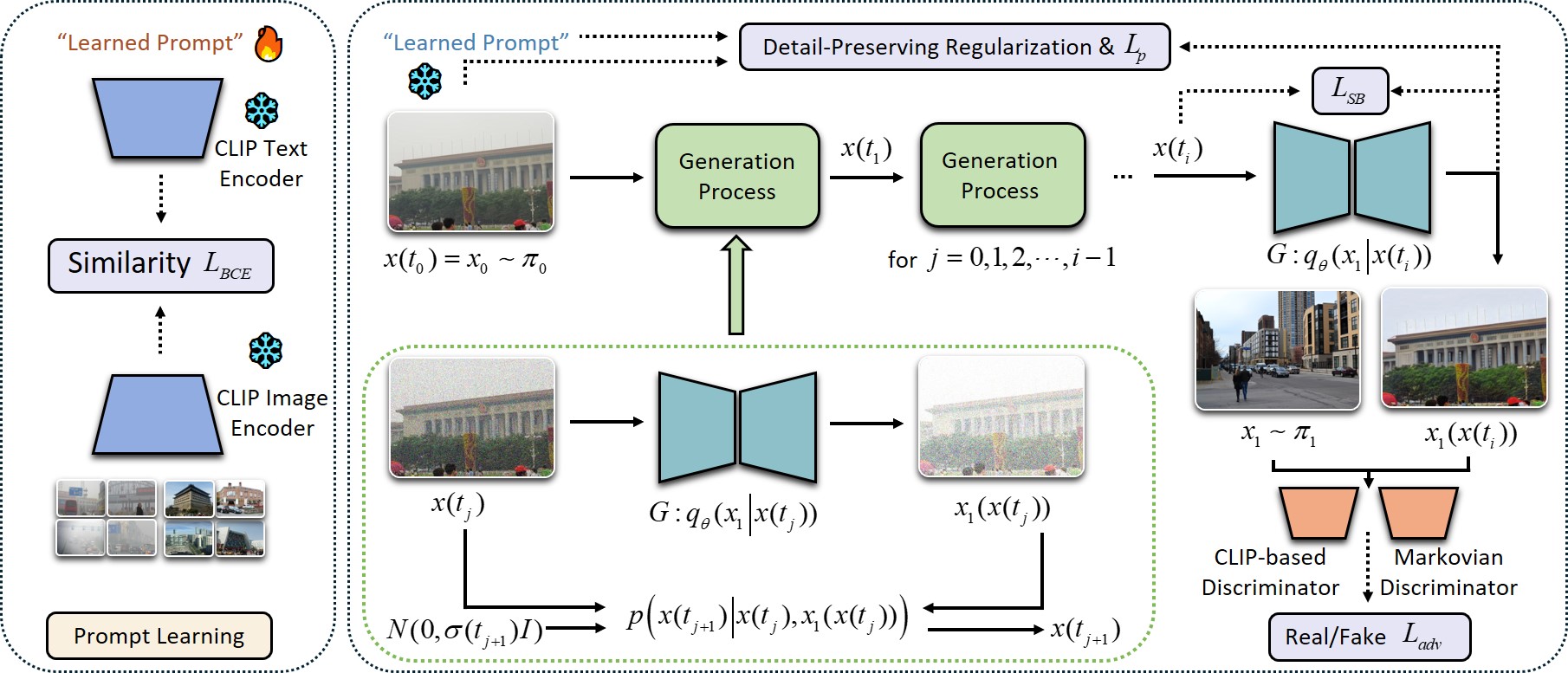} 
\caption{Overview of our method. We first learn a haze-aware prompt using prompt learning. This learned prompt is then used to guide the training of the Schrödinger Bridge (The part within the black dashed box on the right). For SB training, we discretize the unit interval [0, 1] into $N$ intervals with uniform spacing, where $t_j<t_{j+1}$, $t_{0}=0$, and $t_{N}=1$. $x(t_{0})=x_{0}$ denotes the hazy image sampled from $\pi_{0}$. $x_{1}$ denotes the unpaired clear image sampled from $\pi_{1}$. $x_{1}(x(t_{j}))$ denotes the generated dehazed image from intermediate states $x(t_{j})$ using generator $G$. The generation process involves generating $x(t_{j+1})$ from $x(t_j)$ (details are provided within the green dashed box). 
During the training of the SB, we use the hazy image $x_0$ to generate an intermediate state $x(t_i)$ through an iterative generation process with $j=0,1,2,\dots, i-1$.
Next, we feed $x(t_i)$ into the generator $G:q_{\theta}(x_1\mid x(t_{i}))$ to produce a dehazed image $x_1(x(t_i))$.
Finally, we calculate the $L_{adv}$, $L_{SB}$ and $L_{p}$ using the unpaired clear image $x_1$, the generated dehazed image $x_{1}(x(t_{i}))$, the hazy image $x_0$, and the learned prompt. 
Furthermore, we impose a detail-preserving regularization between $x_0$ and $x_{1}(x(t_{i}))$ to enforce structural and detail consistency, ensuring that the dehazed image aligns pixel-by-pixel with the hazy image. In the inference process, we input hazy images directly into generator $G$ to produce dehazed images with different numbers of function evaluations (NFEs).}
\label{figure_2}
\end{figure*}
\subsection{Image Dehazing}
Early prior-based methods focus on obtaining the ASM's parameters, and then restoring clear images by reserving the ASM. For example, DCP \cite{he2010single} proposes a dark channel prior to estimate the atmospheric light and transmission map.
With the development of deep learning, some efforts concentrate on designing networks to estimate the parameters of ASM or directly learn the mapping from hazy to clear images. DehazeNet \cite{cai2016dehazenet} is the pioneer in estimating transmission maps in an end-to-end manner. C2PNet \cite{zheng2023curricular} customizes a curriculum learning to improve dehazing performance. 
Despite the existence of various proposed networks, their dehazing performance on real-world images remains limited since they are trained on synthetic datasets.

To address the bottleneck of paired training, recent studies have focused on unpaired training strategies to improve the model's generalization capabilities.
For example, RefineDNet \cite{zhao2021refinednet} integrates the dark channel prior (DCP) into a two-stage dehazing network for unpaired learning. D4+ \cite{yang2024robust} decomposes the ASM and develops a self-augmented dehazing framework built on CycleGAN \cite{zhu2017unpaired}. UCL-Dehaze \cite{wang2024ucl} proposes an unpaired contrastive learning paradigm for image dehazing. These methods all employ GANs to generate dehazed images. Despite achieving better performance on real-world hazy images, they still fail to produce satisfactory dehazing results due to the limited transmission mapping capability of the generator. 
\subsection{Schrödinger Bridge based Image Translation}

The Schrödinger Bridge \cite{schrodinger1932theorie} problem involves finding the stochastic process that transitions from an initial distribution to a target distribution while adhering to a reference measure.
The Schrödinger Bridge provides the flexibility to select arbitrary distributions as both the initial and terminal distributions, making it highly versatile.
For instance, I$^{2}$SB \cite{liu2023i2sb} proposes the Image-to-Image Schrödinger Bridge, learning nonlinear diffusion processes between two given distributions. 
UNSB \cite{kim2023unpaired} is the first to achieve high-resolution unpaired image translation using adversarial learning. CUNSB-RFIE \cite{dong2024cunsb} introduces a context-aware unpaired neural Schrödinger Bridge for retinal fundus image enhancement.
However, despite these advancements, applying the Schrödinger Bridge to image dehazing remains challenging, since the degradation mechanisms of images captured under hazy conditions are extremely complex.

\section{Preliminary}
\subsection{Schrödinger Bridge}
The Schrödinger Bridge problem involves finding an optimal stochastic process
$\left \{ x(t):t\in [0,1]\right \}$ that connects two probability distributions $\pi_{0}$ and $\pi_{1}$ on $\mathbb{R}^d$, while respecting a reference measure $\mathbb{W}$, typically chosen to be the Wiener measure. Formally, let $\Omega$  represent the path space and $\mathcal{P}\left ( \Omega\right )$ denote the probability measure on $\Omega$. The Schrödinger Bridge problem can be mathematically formulated as:
\begin{equation}
\label{eq_2}
\mathbb{Q}^{SB} = \underset{\mathbb{Q} \in \mathcal{P}(\Omega)}{\arg\min}\ D_{KL}(\mathbb{Q} \parallel \mathbb{W}) \quad
\text{s.t.} \quad
\mathbb{Q}_0 = \pi_0,\ \mathbb{Q}_1 = \pi_1,
\end{equation}
where $\mathbb{Q}^{SB}$ denotes the Schrödinger Bridge solution, and $\mathbb{Q}_{t}$  corresponds to the marginal distribution of $\mathbb{Q}$ at time $t$. $D_{KL}(\mathbb{Q} \parallel \mathbb{W})$ represents the Kullback-Leibler divergence.
This formulation establishes a probabilistic coupling that enables bidirectional sampling between the distributions. Specifically, given samples $x_{0}\sim \pi_{0}$, the bridge $\mathbb{Q}^{SB}$ facilitates the generation of corresponding samples $x_{1}\sim \pi_{1}$ through the interpolating process. 
The reverse direction is equally attainable through time reversal.
\subsection{Static Schrödinger Bridge}
The Schrödinger Bridge enables the direct learning of optimal transport between arbitrary distributions. However, the inherent temporal dependencies in the full process make the direct optimization of $\mathbb{Q}^{SB}$ challenging.
The Static Schrödinger Bridge (SSB) offers a solution by simplifying the Schrödinger Bridge problem into a more tractable formulation. This reduction establishes an equivalence between the Schrödinger Bridge $\mathbb{Q}^{SB}$ and its static formulation ${\mathbb{Q}}_{01}^{SB}$, which characterizes the joint distribution of the initial distribution $\pi_{0}$ and the terminal distribution $\pi_{1}$. 
Let $\Pi(\pi_0,\pi_1)$ denote the space of joint distributions with marginals $\pi_{0}$ and $\pi_{1}$.
The SSB problem can be formulated as an entropy-regularized optimal transport problem:
\begin{equation}
\label{eq_3}
\mathbb{Q}_{01}^{\text{SB}} = \underset{\gamma \in \Pi(\pi_0,\pi_1)}{\arg\min}\ \mathbb{E}_{(x_0,x_1) \sim \gamma} \left[ \| x_0 - x_1 \|^2 \right] - 2\tau H(\gamma),
\end{equation}
where $H(\gamma)$ denotes the entropy, and $\tau$ controls the amount of randomness in the trajectory. Under this formulation, the marginal distribution $x(t):t\in [0,1]$ can be obtained through conditional Gaussian distributions:
\begin{equation}
\label{eq_4}
p{(x(t)\mid x_{0},x_{1})} \sim \mathcal{N}(x(t);tx_{1}+(1-t)x_{0},t(1-t)\tau I),
\end{equation}
where $I$ is the identity matrix. A crucial self-similarity property emerges in $\mathbb{Q}_{01}^{\text{SB}}$, as demonstrated in UNSB \cite{kim2023unpaired}. 
For any sub-interval $[t_{\alpha},t_{\beta}]\subseteq [0,1]$, the restricted process ${Q}_{t_{\alpha}t_{\beta}}^{SB}$ remains a Schrödinger Bridge, which can be formulated as conditional Gaussian distributions:
\begin{equation}
\begin{aligned}
\label{eq_5}
p{(x(t)\mid x(t_{\alpha}),x(t_{\beta}))}\sim \mathcal{N}(x(t); \mu(t), \sigma(t)),  \\ 
\mu(t) = s(t)x(t_\beta)+(1-s(t))x(t_{\alpha}), \\
\sigma(t) = s(t)(1-s(t))\tau (t_\beta-t_\alpha)I,
\end{aligned}
\end{equation}
where $s(t) := (t-t_{\alpha})/(t_{\beta}-t_{\alpha})$. This enables recursive interpolation between intermediate states.
\section{Method}

As shown in Figure \ref{figure_2}, We propose a novel unpaired dehazing framework, DehazeSB, based on the Schrödinger Bridge. It comprises three components: prompt learning, Schrödinger Bridge, and detail-preserving regularization.
\subsection{Prompt Learning}
\begin{figure}[t]
\centering
\includegraphics[width=0.8\linewidth]{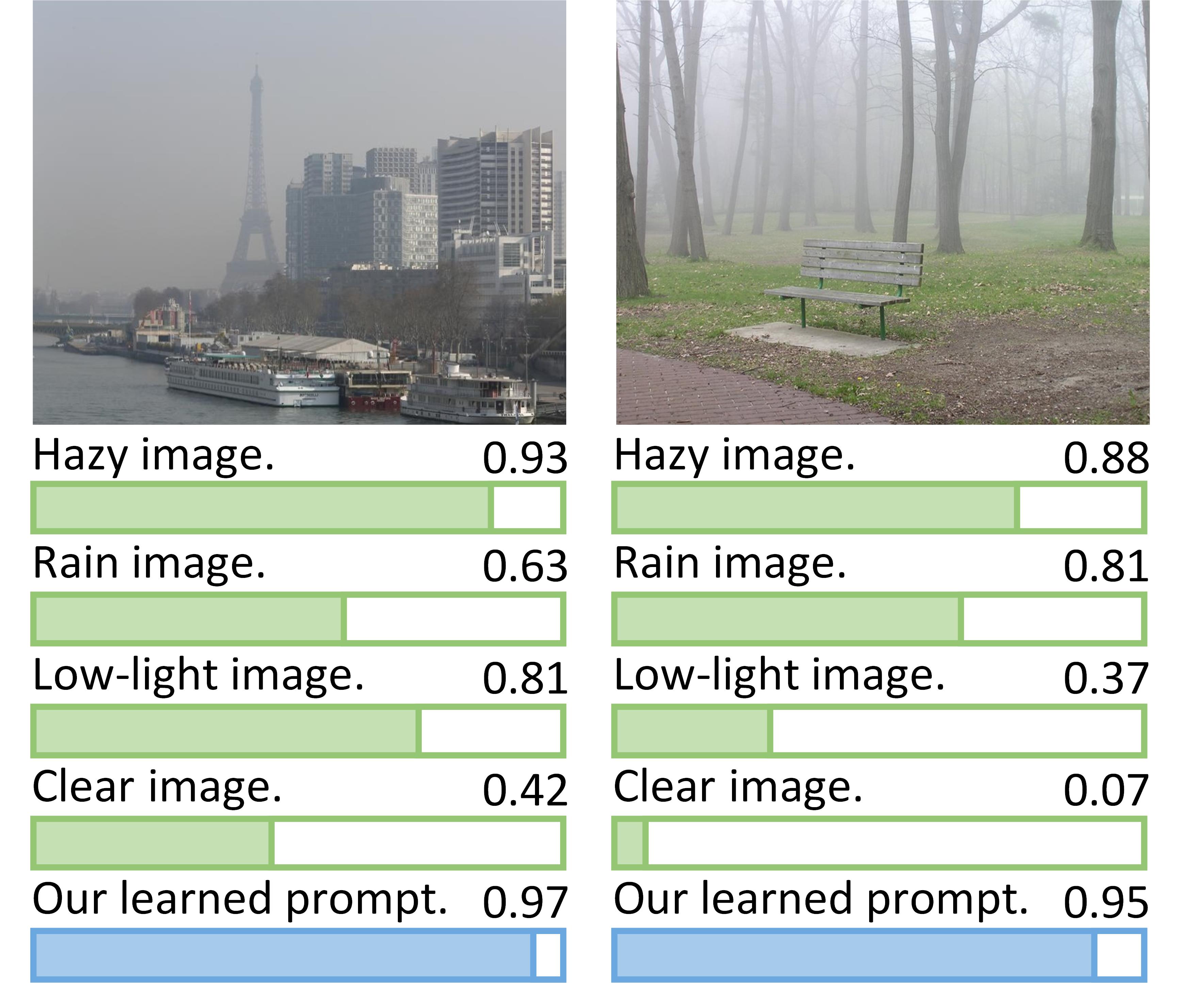}
\caption{CLIP scores of different prompts. 
We observe that appropriate prompts are effective in distinguishing between hazy and clear images (for instance, the prompt ``\textit{hazy image}" receives a higher score on hazy images). However, due to the complex degradation characteristics of real-world hazy images, simplistic prompts like ``\textit{hazy image}" fail to accurately capture these nuances. In contrast, our learned haze-aware prompt demonstrates superior effectiveness, achieving the highest scores.}
\label{figure_3}
\end{figure}
Inspired by CLIP's \cite{clip,liang2023iterative,morawski2024} demonstrated capability for cross-modal alignment, we observe that its discriminative power extends to distinguishing hazy and clear images when provided with textual descriptors (as validated in Figure \ref{figure_3}).  
This insight motivates us to leverage prompts to facilitate dehazing. 
However, we have empirically observed that generic prompts such as ``\textit{hazy image}'' yield sub-optimal results. 
This limitation arises from the wide variation in haze concentration and distribution found in real-world hazy images, which simplistic prompts fail to adequately capture.

To address this problem, we propose a novel prompt learning that exclusively optimizes haze-aware prompts.
Specifically, we randomly initialize a learnable textual prompt $T_{hazy}$. Subsequently, we leverage the pre-trained CLIP's image encoder $\Phi_{image}$ and text encoder $\Phi_{text}$ to separately extract features from hazy images $I_{hazy}$, clear images $I_{clear}$, and initialized haze-aware prompt $T_{hazy}$. We train this prompt using a binary cross-entropy (BCE) loss:
\begin{equation}
\label{eq_6}
L_{BCE}=-(y\log{\hat{y}}+(1-y)\log(1-\hat{y})),
\end{equation}
\begin{equation}
\label{eq_7}
\hat{y}=\frac{e^{cos((\Phi _{image}(I),\Phi _{text}(T_{hazy})))}}{\textstyle\sum_{i\in \left \{ hazy,clear\right \}}e^{cos((\Phi _{image}(I_{i}),\Phi _{text}(T_{hazy})))}},
\end{equation}
where $I \in \{I_{hazy}, I_{clear}\}$, and $y$ is the label of $I$. 0 is for negative sample $I_{clear}$ and 1 is for positive sample $I_{hazy}$.
The learned haze-aware prompt subsequently guides the training of the Schrödinger Bridge. Specifically, given a generated dehazed image $I_{dehaze}$ (denoted by $x_1(x(t_{i}))$ in Figure \ref{figure_2}) and its hazy counterpart $I_{hazy}$ (denoted by $x_{0}$), we guide the dehazing by pushing the learned haze-aware prompt away from dehazed images. This loss function can also be expressed as Eq. \ref{eq_6}, where $\hat{y}$ is formulated as:
\begin{equation}
\label{eq_9}
\hat{y}=\frac{e^{cos((\Phi _{image}(I),\Phi _{text}(T_{hazy})))}}{\textstyle\sum_{i\in \left \{ hazy,dehaze\right \}}e^{cos((\Phi _{image}(I_{i}),\Phi _{text}(T_{hazy})))}},
\end{equation}
where $I \in \{I_{hazy}, I_{dehaze}\}$. Specifically, 0 is for negative sample $I_{dehaze}$ and 1 is for positive sample $I_{hazy}$. 
\subsection{Schrödinger Bridge}
For SB training, we discretize the unit interval $[0, 1]$ into $N$ intervals with uniform spacing ($N=5$ in our framework).
According to the Markov chain decomposition, we can simulate the SB between hazy and clear images as:
\begin{equation}
\label{eq_10}
p(\{x(t_n)\}) = p(x(t_N) | x(t_{N-1}))\cdots p(x({t_1}) | x({t_0})) p(x({t_0})),
\end{equation}
where $t_i<t_{i+1}$, $t_0=0$, and $t_N=1$. $\{x({t_n})\}$ represents a partition ${\{x(t_i)\}}_{i=0}^{N}$ of the unit interval $\left [ 0,1\right ]$. 
Given the initial hazy image distribution $p(x(t_{0}))$, learning the transition probabilities $p(x(t_{i+1}) | x(t_{i}))$ enables recursive sampling to approximate the target distribution $p(x(t_{N}))$, where $p(x(t_{0}))$ and $p(x(t_{N}))$ are also refered to as $p(x_0)$ and $p(x_1)$.  
Therefore, we need to design a network with parameters $\theta_{i}$ as the generator $G:q_{\theta _{i}}(x(t_{i+1})\mid x(t_{i}))$ to modulate $p(x(t_{i+1}) \mid x(t_{i}))$, allowing us to sample from the initial distribution and obtain the target distribution. 

However, it is impractical to devise a separate generator $G$ for each time step. Following the UNSB \cite{kim2023unpaired}, we devise an elaborated generator $G:q_{\theta}(x(t_N)\mid x(t_{i}))$ conditioned on time step $t_{i}$, which can also be denoted as $G:q_{\theta}(x_1\mid x(t_{i}))$. This generator directly predicts the target $x_1$ from any intermediate state $x(t_{i})$. In doing so, the optimization objective can be expressed as:
\begin{equation}
\begin{aligned}
\label{eq_11}
L_{SB}:= \mathbb{E}_{q_{\theta}(x(t_i), x_1)} \left[ \| x(t_i) - x_1 \|^2 \right] \\
- 2\tau (1 - t_i) H(q_{\theta}(x(t_i), x_1)) \\
\text{s.t.}\quad L_{\text{adv}}:= D_{\text{KL}}(q_{\theta}(x_1) \| p(x_1)) = 0.
\end{aligned}
\end{equation}
Using Lagrange multipliers, we can transform the optimization objective into a loss function:
\begin{equation}
\label{eq_12}
L:=L_{adv}+\lambda_{SB}L_{SB}.
\end{equation}
The training pipeline is illustrated in Figure \ref{figure_2}.
Starting from an initial sample $x(t_{0})=x_{0}\sim \pi_{0}$, we progressively generate intermediate states $\left \{x(t_{j}) \right \} _{j=1}^i$ through a Markov chain process. 
After obtaining $x(t_{i})$, we use the generator $G:q_{\theta}(x_{1}\mid x(t_{i}))$ to produce the dehazed image $x_1(x(t_{i}))$. We then calculate the $L_{SB}$ and $L_{adv}$ in Eq.\ref{eq_11} using the pairs $(x(t_{i}),x_{1}(x(t_{i})))$ and $(x_{1}({x(t_{i}})),x_{1})$. Note that we randomly choose different time steps $t_{i}$ to optimize during the training, thus our model can generate target images with an arbitrary number of function evaluations (NFEs).
For the generation process of $x(t_{j+1})$ at each step $j=0,1,2,\cdots,i-1$, we first predict the target domain sample $x_1(x(t_{j}))$ using the generator $G:q_{\theta}(x_{1}\mid x(t_{j}))$. Subsequently we generate $x(t_{j+1})$ through the Eq. \ref{eq_5}, which can be expressed as:
\begin{equation}
\begin{aligned}
\label{eq_13}
p{(x(t_{j+1})\mid x(t_{j}),x_{1}(x(t_{j})))}\sim \\
\mathcal{N}(x(t_{j+1}); \mu(t_{j+1}), \sigma(t_{j+1})),  \\ 
\mu(t_{j+1}) = s(t_{j+1})x_{1}(x(t_{j}))+(1-s(t_{j+1}))x(t_{j}), \\
\sigma(t_{j+1}) = s(t_{j+1})(1-s(t_{j+1}))\tau (1-t_{j})I.
\end{aligned}
\end{equation}
To align the distribution of generated images $q_{\theta}(x_1)$ with the target distribution $p(x_1)$, we replace the KL divergence in Eq.\ref{eq_11} with a dual-discriminator adversarial strategy.
Specifically, We first employ a Markovian discriminator which distinguishes samples on the patch level since it is effective at capturing details and textures in local characteristics. 
To complement patch-level analysis, we introduce a CLIP-based discriminator \cite{kumari2022ensembling} that assesses overall structural coherence and scene consistency. This mitigates limitations of purely local evaluation, which can overlook implausible global arrangements. 
By combining these discriminators, our method simultaneously enhances local realism and global plausibility. Finally, We compute 
$H(q_{\theta}(x(t_i), x_1))$ in a manner similar to the approach used in UNSB \cite{kim2023unpaired}, as detailed in our supplementary materials.

\subsection{Detail-Preserving Regularization}

Another critical challenge in unpaired image dehazing is detail preservation. The generated images must not only display high visual quality but also maintain fidelity in details and textures compared to the original images. Although CycleGAN \cite{zhu2017unpaired} is a classic solution for this problem, it necessitates the training of additional generators and discriminators, leading to high computational demands. To overcome this limitation, we introduce a detail-preserving regularization that includes PatchNCE regularization, physical prior regularization, and high-frequency detail regularization.

\subsubsection{PatchNCE Regularization}

Following the CUT \cite{park2020cut}, we employ PatchNCE regularization to enhance the generator's ability to preserve content consistency during image dehazing.
Specifically, PatchNCE regularization leverages contrastive learning to enforce similarity between corresponding patches in the input and output. This helps the generator learn to maintain important structural details and textures, thereby improving the overall quality of the dehazed images. For more detailed information, please refer to our supplementary materials.

\subsubsection{Physical Prior Regularization}

\begin{figure*}[t]
\centering
\includegraphics[width=0.8\textwidth]{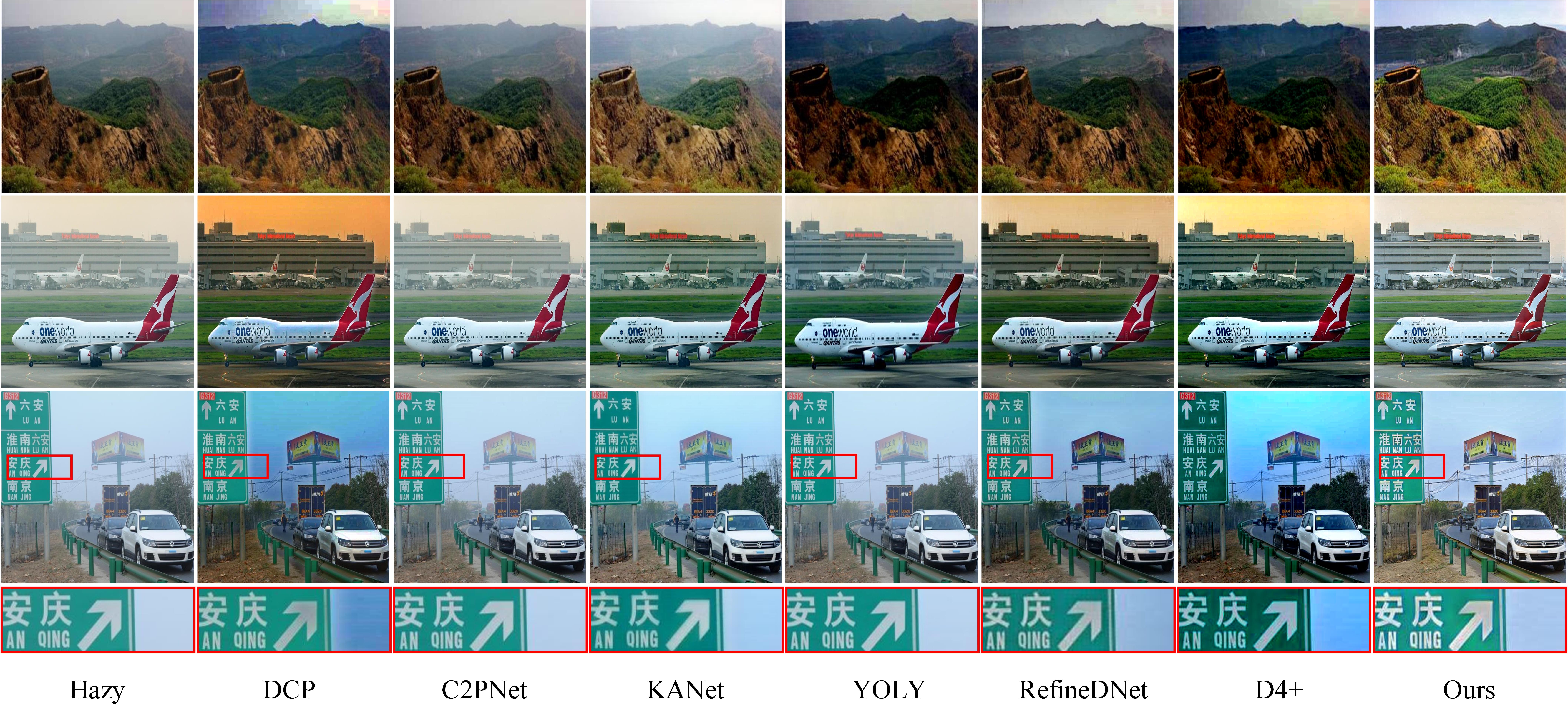} 
\caption{Visual comparison of samples from RTTS and Haze2020. Our method can effectively remove haze and generate high-quality images with natural color and realistic contrast. \textbf{More than 40 visual results are presented in our supplementary materials.}}
\label{figure_rtts}
\end{figure*}
\begin{table*}[t]
\centering
\small
\begin{tabular}{cccccccccc}
\toprule
  & & \multicolumn{4}{c}{RTTS} & \multicolumn{4}{c}{Haze2020} \\ \midrule
 Type & Method & FID$\downarrow$  & NIQE$\downarrow$ & MUSIQ$\uparrow$ & MANIQA $\uparrow$ & FID$\downarrow$ & NIQE$\downarrow$ & MUSIQ$\uparrow$ & MANIQA$\uparrow$\\ \midrule
Prior-based &  DCP \cite{he2010single} & 73.017 & 4.271 & 52.935 & 0.119 & 91.948 & 3.827 & 54.375 & 0.121 \\ 
Paired & MBFormer \cite{qiu2023mb} & 66.023 & 4.874 & 53.576 & 0.139 & 82.596 & 4.128 & 55.550 & \underline{0.141}\\
Paired & C2PNet \cite{zheng2023curricular}& 66.117 & 5.037 & 53.960 & 0.142 & 83.959 & 4.206 & 54.565 & 0.138\\
Paired & KANet \cite{feng2024kanet}& 64.963 & 4.339 & 54.513 & 0.110 & 84.888 & \textbf{3.742} & 56.411 & 0.127\\
Paired & DEANet \cite{chen2024dea} & 66.355 & 4.805 & 53.628 & 0.132 & 84.197 & 3.976 & 54.757 & 0.135 \\
Paired & Diff-Plugin \cite{liu2024diff} & 65.787 & 5.334 & 50.740 & 0.095 & 80.965 & 4.407 & 52.752 & 0.111\\
Paired & OneRestore \cite{guo2024onerestore} & 65.934 & 5.321& 51.380 & 0.114 & 84.276 & 4.365 & 52.066 & 0.116\\
Paired & SGDN \cite{fang2024sgdn} & 78.018 & 5.716 & 44.858 & 0.081 & 89.061 & 5.914 & 46.326 & 0.094\\
Unpaired & CUT \cite{park2020cut} & 98.771 & 3.938 & \underline{59.798} & \underline{0.146} & 75.625 & 4.261 & \underline{58.668} & 0.135\\
Unpaired & UNSB \cite{dong2024cunsb} & \underline{56.668} & \underline{3.834} & 55.048  & 0.132 & \underline{72.812} & 4.137 & 55.384 & 0.132\\
Unpaired & YOLY \cite{li2021yoly} & 69.996 & 4.553 & 51.874 & 0.129& 85.648 & 3.937 & 54.164 & 0.138\\
Unpaired & RefineDNet \cite{zhao2021refinednet} & 65.076 & 4.012 & 56.117 & 0.108 & 88.229 & 3.972 & 54.779 & 0.122\\
Unpaired & D4 \cite{yang2022self} & 69.400 & 4.637 & 58.445 & 0.139 & 87.536 & 3.971 & 53.458 & 0.119\\
Unpaired & D4+ \cite{yang2024robust} & 67.687 & 4.274 & 53.599 & 0.131 & 87.101 & 3.865 & 53.612 & 0.124 \\
Unpaired & Ours & \textbf{53.120} & \textbf{3.760} & \textbf{60.114} & \textbf{0.153} & \textbf{69.796} & \underline{3.743} & \textbf{59.256} & \textbf{0.150} \\ \bottomrule
\end{tabular}
\caption{Quantitative results on RTTS and Haze2020. The best results are denoted in \textbf{bold}, and the second-best results are \underline{underlined}.}
\label{table_1}
\end{table*}
Previous studies \cite{zhao2021refinednet,yang2022self,chen2021psd} have demonstrated that using physical priors can enhance real-world image dehazing.
Moreover, they enable us to preserve as much detail as possible. Specifically, we process the hazy image $I$ (also referred to as $x_{0}$) using  Dark Channel Prior (DCP) \cite{he2010single}, obtaining the atmospheric light $A$, coarse transmission map $t$, and dehazed image $J_{dcp}$. We then feed $t$ into a UNet \cite{ronneberger2015u} to obtain a refined transmission map $t_{ref}$. Finally, we reconstruct the hazy image $I_{phy}$ according to the Eq. \ref{eq_1} and formulate the physical prior regularization as follows:
\begin{equation}
\label{eq_14}
L_{phy}=L_{rec}(I,I_{phy}),
\end{equation}
where ${I}_{phy}=J{t}_{ref}+A(1-{t}_{ref})$ refers to the reconstructed hazy image derived using the ASM, and $J$ denotes the generated dehazed image $x_{1}(x(t_{i}))$. ${L}_{rec}$ is the combined distance of ${L}_{1}$ and LPIPS. \cite{zhang2018unreasonable}.

\subsubsection{High-Frequency Detail Regularization}

To enhance structural fidelity in the restored images while preserving high-frequency details, we introduce a high-frequency detail regularization that includes Discrete Fourier Transform (DFT) loss, Structural Similarity Index Measure (SSIM) loss, and Sobel Gradient loss.
We provide more detailed information in our supplementary materials. 

The final loss function can be formulated as:
\begin{equation}
\begin{aligned}
\label{eq_15}
L=L_{adv} + \lambda_{SB}L_{SB} + \lambda_{p}L_{p}+ \\ 
\lambda_{NCE}L_{NCE} + \lambda_{phy}L_{phy}+\lambda_{hfd}L_{hfd},
\end{aligned}
\end{equation}
where $L_{NCE}$, $L_{phy}$, and $L_{hfd}$ denote the PatchNCE, physical prior, and high-frequency detail regularization loss. Each term is weighted by its corresponding coefficients $\lambda$: 1, 1, 1, 0.5, and 0.5 respectively. 
\section{Exprements and Discussion}

\begin{figure*}[t]
\centering
\includegraphics[width=0.8\textwidth]{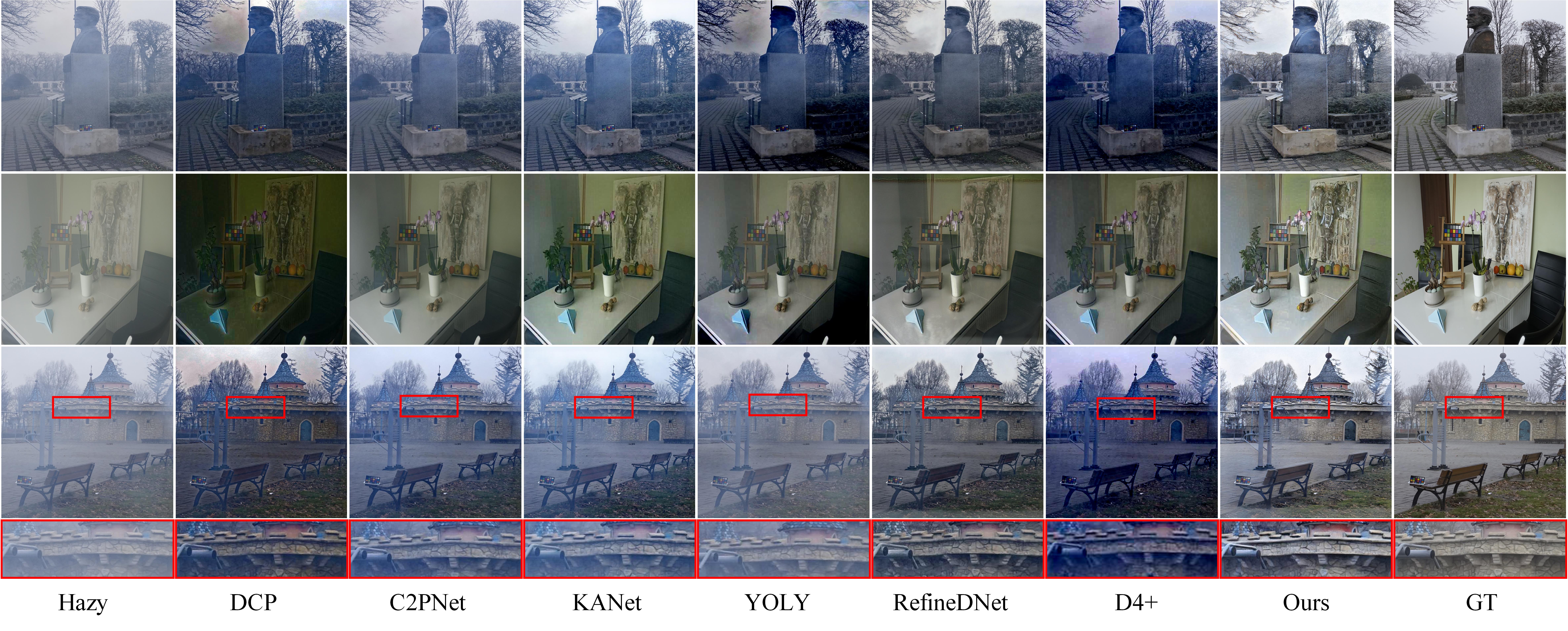} 
\caption{Visual comparison of samples from OHAZE and IHAZE.}
\label{figure_ohaze}
\end{figure*}
\begin{table}[t]
\centering
\small
\begin{tabular}{ccccc}
\toprule
Method & PSNR$\uparrow$  & SSIM$\uparrow$& LPIPS$\downarrow$&VSI$\uparrow$ \\ \midrule
DCP & 17.005 & \underline{0.819} & \underline{0.242} & 0.945 \\  
MBFormer & 15.243 & 0.672 & 0.339 & 0.911 \\
C2PNet & 18.014 & 0.708 & 0.317 & 0.929 \\
KANet & 17.713 & 0.814 & 0.265 & 0.939 \\
DEANet & 14.687 & 0.691 & 0.440 & 0.880 \\
Diff-Plugin & 16.470 & 0.526 & 0.364 & 0.920 \\
OneRestore & 17.458 & 0.740 & 0.282 & 0.924 \\
SGDN & 18.163 & 0.682 & 0.381 & 0.953 \\
CUT & 15.927 & 0.713 & 0.341 & 0.936 \\
UNSB & 15.018 & 0.648 & 0.392 & 0.940 \\
YOLY & 14.974 & 0.406 & 0.425 & 0.914 \\
RefineDNet & \underline{18.693} & 0.755 & 0.260 & \underline{0.954} \\
D4 & 16.767 & 0.690 & 0.290 & 0.939 \\
D4+ & 17.336 & 0.677 & 0.301 & 0.947 \\
Ours & \textbf{18.829} & \textbf{0.838} & \textbf{0.223} & \textbf{0.961} \\ \bottomrule
\end{tabular}
\caption{Quantitative results on OHAZE.}
\label{table_2}
\end{table}
Our method enables direct training on unpaired real-world hazy and clear images. Specifically, we collect over 7,000 real hazy images from the RESIDE \cite{li2018benchmarking} for our training set of hazy images. Additionally, we gather over 10,000 clear images from the SOTS (a subset of RESIDE) and ADE20K \cite{zhou2019semantic} to serve as our training set of clear images.
To comprehensively evaluate our method, we conduct extensive quantitative and qualitative comparisons using multiple real-world datasets: RTTS, URHI, Haze2020, OHAZE, IHAZE, and Fattal's dataset. Among them, RTTS and URHI are subsets of RESIDE, each containing over 4,000 hazy images. Haze2020, introduced by DA \cite{shao2020domain}, includes more than 1,000 real-world images. OHAZE \cite{ancuti2018haze} and IHAZE \cite{ihaze} contain 45 and 55 pairs of hazy images, respectively, generated using a haze machine. Fattal's dataset \cite{fattal2014dehazing} comprises over 30 real-world hazy images.
More training details and parameters are in the supplementary materials.

We compare our method against several state-of-the-art methods. Among them, DCP is a prior-based dehazing method. CUT and UNSB are originally proposed for unpaired image translation.
For images with available ground truth (GT), we evaluate them using full-reference metrics such as PSNR, SSIM \cite{wang2004image}, LPIPS, and VSI \cite{zhang2014vsi}. For images without GT, we use no-reference metrics, including FID \cite{heusel2017gans}, NIQE \cite{mittal2012making}, MUSIQ \cite{ke2021musiq}, and MANIQA \cite{yang2022maniqa}.
\subsection{Results on RTTS and Haze2020}

We conduct qualitative comparisons on RTTS and Haze2020, and zoom in on local details within partial images. As shown in Figure \ref{figure_rtts}, DCP effectively processes images but tends to over-enhance results due to the limitations of hand-crafted priors.
C2PNet struggles with images because it is trained on synthetic datasets, which have a domain gap with real-world data. KANet shows promising performance by employing an advanced data synthesis strategy. However, KANet still falls short of fully removing haze due to the inherent limitations of paired training.
Unpaired dehazing methods like YOLY, RefineDNet, and D4+ can dehaze images to some extent. However, these methods often produce sub-optimal results due to the limited transport mapping capabilities of GAN generators, resulting in artifacts and insufficient detail.
In contrast, our proposed method leverages the Schrödinger Bridge to directly learn optimal transport mappings in an unpaired training framework. This approach generates more realistic and natural images with rich details, textures, and high contrast.

Table \ref{table_1} presents a quantitative comparison of Haze2020 and RTTS. Our method achieves the best quantitative results on RTTS and excels in several metrics on Haze2020. For NIQE, our approach ranks second, slightly behind KANet. This discrepancy likely results from the introduction of physical prior regularization in our method.
We introduce physical prior regularization to ensure our method effectively removes haze while preserving the original image's fine details and textures. Although this regularization slightly compromised NIQE scores, we believe the overall improvement in image quality justifies this trade-off.


\subsection{Results on OHAZE and IHAZE}
Figure \ref{figure_ohaze} and Table \ref{table_2} present results on OHAZE and IHAZE. Notably, we do not retrain our model on these datasets but directly evaluate them to demonstrate the generalization capability of our method.
The experimental results indicate that our method outperforms other approaches, achieving superior metrics while also producing visually pleasing dehazed images.
The results for URHI and Fattal's dataset are provided in the supplementary materials.

\subsection{Ablation Study}
\begin{figure}[t]
\centering
\includegraphics[width=0.85\linewidth]{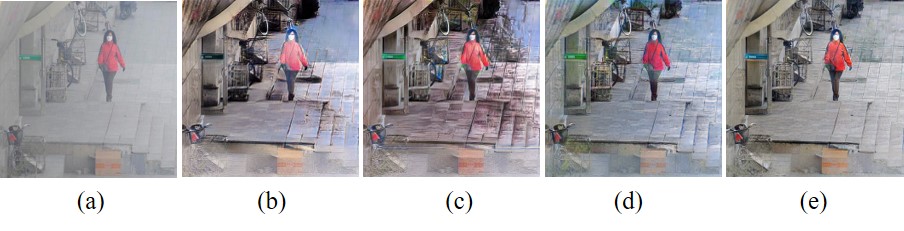}
\caption{Ablation study of each component. (a) Hazy image. (b) Backbone. (c) Backbone + PL. (d) Backbone + DPR. (e) Ours.}
\label{ablation_main}
\end{figure}
\begin{table}[t]
\centering
\small
\begin{tabular}{ccccc}
\toprule
Backbone & \checkmark & \checkmark & \checkmark & \checkmark \\
PL & & \checkmark & & \checkmark \\
DPR & & & \checkmark & \checkmark \\ \midrule
FID $\downarrow$ & 80.260 & 73.968 & 75.192 & \textbf{69.796} \\
NIQE $\downarrow$ & 3.712 & \textbf{3.690} & 3.981 & 3.743 \\
MUSIQ $\uparrow$ & 54.414 & 59.109 & 56.550 & \textbf{59.256} \\
MANIQA $\uparrow$ & 0.105 & 0.144 & 0.133 & \textbf{0.150} \\ \midrule
PSNR $\uparrow$ & 17.410 & 17.445 & 18.687 & \textbf{18.829} \\
SSIM $\uparrow$ & 0.777 & 0.796 & \textbf{0.839} & 0.838 \\
LPIPS $\downarrow$ & 0.282 & 0.298 & 0.253 & \textbf{0.223} \\
VSI $\uparrow$ & 0.939 & 0.937 & 0.952 & \textbf{0.961} \\ \bottomrule
\end{tabular}
\caption{Ablation study of each component.}
\label{table_3}
\end{table}

We conduct a series of ablation studies on Haze2020 and OHAZE to validate the effectiveness of each component, including the Schrödinger Bridge (SB), Prompt Learning (PL), and Detail-Preserving Regularization (DPR).
Note that our method employs unpaired training, which does not utilize ground truth (GT). Therefore, PatchNCE regularization is essential to prevent gradient explosion. In the following ablation study, our backbone includes PatchNCE regularization by default.
We construct the following variants: \textbf{Backbone}: Removing both PL and DPR. \textbf{Backbone + PL} : Removing DPR. \textbf{Backbone + DPR}: Removing PL. \textbf{Ours}: Our method. 
As shown in Table \ref{table_3} and Figure \ref{ablation_main}, our proposed method achieves the best performance in both visual appeal and most metrics, striking an effective balance between perceptual quality and distortion. This validates that each component plays a critical role in improving dehazing performance.


\subsubsection{Ablation of Schrödinger Bridge}

\begin{figure}[t]
\centering
\includegraphics[width=0.7\linewidth]{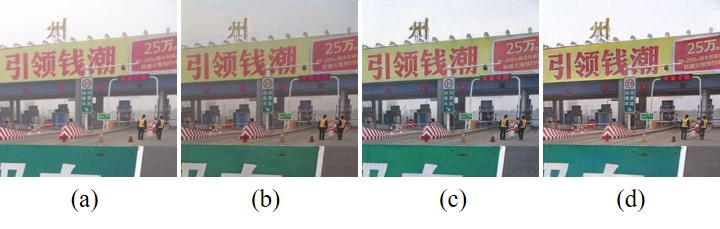}
\caption{Ablation study of discriminators. (a) Hazy image. (b) Remove MD. (c) Remove CD. (d) Ours.}
\label{ablation_gan}
\end{figure}

We perform an ablation study to evaluate how discriminators affect the results within our framework. As illustrated in Figure \ref{ablation_gan}, by employing both a Markovian Discriminator (MD) and a CLIP-based Discriminator (CD), we can simultaneously preserve the local details and ensure the global quality of the dehazed images, resulting in high-quality results. 
Moreover, the MD plays a crucial role in our method due to our use of PatchNCE regularization, which minimizes the distance between a patch in the dehazed image and its corresponding patch in the hazy image at the same location. Without the MD to evaluate images at the patch level, the method will fail to effectively remove haze.

\begin{table}[t]
\centering
\small
\begin{tabular}{ccccc}
\toprule
Backbone & \checkmark & \checkmark & \checkmark & \checkmark $ $\\
PPR & & \checkmark & & \checkmark \\
HFDR & & & \checkmark & \checkmark \\ \midrule
FID $\downarrow$ & 73.968 & 74.372 & 73.727 & \textbf{69.796} \\
NIQE $\downarrow$ & 3.690 & 3.753 & \textbf{3.655} & 3.743 \\
MUSIQ $\uparrow$ & 59.109 & 55.770 & 58.526 & \textbf{59.256} \\
MANIQA $\uparrow$ & 0.144 & 0.143 & 0.142 & \textbf{0.150} \\ \midrule
PSNR $\uparrow$ & 17.445 & 18.085 & 17.534 & \textbf{18.829} \\
SSIM $\uparrow$ & 0.796 & 0.812 & 0.805 & \textbf{0.838} \\
LPIPS $\downarrow$ & 0.298 & 0.290 & 0.267 & \textbf{0.223} \\
VSI $\uparrow$ & 0.937 & 0.940 & 0.940 & \textbf{0.961} \\ \bottomrule
\end{tabular}
\caption{Ablation study of detail-preservation regularization.}
\label{table_5}
\end{table}

Additionally, we investigate the impact of the number of intervals used in our method, with detailed information provided in the supplementary materials.



\subsubsection{Ablation of detail-preserving regularization}
We validate the effectiveness of detail-preservation regularization, which includes physical prior regularization (PPR) and high-frequency detail regularization (HFDR). Table \ref{table_5} demonstrates that both types of regularization contribute to improved performance. We also analyze the impact of varying weights of them in our supplementary materials.

\section{Conclusion and Discussion}
In this paper, we apply the Schrödinger Bridge in real-world image dehazing with unpaired training, named DehazeSB. By learning the optimal transport mappings from hazy to clear image distributions and enforcing detail-preserving regularization, we achieve high-quality dehazing.
Furthermore, inspired by the observation that pre-trained CLIP models can accurately distinguish between hazy and clear images, we introduce a novel prompt learning to enhance dehazing. Extensive experiments demonstrate its effectiveness in producing high-quality dehazed results.

While our Schrödinger Bridge-based approach shows significant superiority in unpaired image dehazing, it has some limitations. First, it depends on a learned haze-aware prompt, making it sensitive to the quality and diversity of the training data. If the training set lacks variability in haze conditions, the model's generalization may suffer. Second, although our detail-preserving improves fidelity in textures and structures, it struggles with highly complex or fine-grained details, especially in images with dense haze.
{
    \small
    \bibliographystyle{ieeenat_fullname}
    \bibliography{main}

\begin{thebibliography}{51}
\providecommand{\natexlab}[1]{#1}
\providecommand{\url}[1]{\texttt{#1}}
\expandafter\ifx\csname urlstyle\endcsname\relax
  \providecommand{\doi}[1]{doi: #1}\else
  \providecommand{\doi}{doi: \begingroup \urlstyle{rm}\Url}\fi

\bibitem[Ancuti et~al.(2018{\natexlab{a}})Ancuti, Ancuti, Timofte, and
  De~Vleeschouwer]{ihaze}
Cosmin Ancuti, Codruta~O Ancuti, Radu Timofte, and Christophe De~Vleeschouwer.
\newblock I-haze: A dehazing benchmark with real hazy and haze-free indoor
  images.
\newblock In \emph{Advanced Concepts for Intelligent Vision Systems: 19th
  International Conference, ACIVS 2018, Poitiers, France, September 24--27,
  2018, Proceedings 19}, pages 620--631. Springer, 2018{\natexlab{a}}.

\bibitem[Ancuti et~al.(2018{\natexlab{b}})Ancuti, Ancuti, Timofte, and
  De~Vleeschouwer]{ancuti2018haze}
Codruta~O Ancuti, Cosmin Ancuti, Radu Timofte, and Christophe De~Vleeschouwer.
\newblock O-haze: a dehazing benchmark with real hazy and haze-free outdoor
  images.
\newblock In \emph{Proceedings of the IEEE conference on computer vision and
  pattern recognition workshops}, pages 754--762, 2018{\natexlab{b}}.

\bibitem[Berman et~al.(2016)Berman, Avidan, et~al.]{berman2016non}
Dana Berman, Shai Avidan, et~al.
\newblock Non-local image dehazing.
\newblock In \emph{Proceedings of the IEEE conference on computer vision and
  pattern recognition}, pages 1674--1682, 2016.

\bibitem[Cai et~al.(2016)Cai, Xu, Jia, Qing, and Tao]{cai2016dehazenet}
Bolun Cai, Xiangmin Xu, Kui Jia, Chunmei Qing, and Dacheng Tao.
\newblock Dehazenet: An end-to-end system for single image haze removal.
\newblock \emph{IEEE transactions on image processing}, 25\penalty0
  (11):\penalty0 5187--5198, 2016.

\bibitem[Chen et~al.(2021)Chen, Wang, Yang, and Liu]{chen2021psd}
Zeyuan Chen, Yangchao Wang, Yang Yang, and Dong Liu.
\newblock Psd: Principled synthetic-to-real dehazing guided by physical priors.
\newblock In \emph{Proceedings of the IEEE/CVF conference on computer vision
  and pattern recognition}, pages 7180--7189, 2021.

\bibitem[Chen et~al.(2024)Chen, He, and Lu]{chen2024dea}
Zixuan Chen, Zewei He, and Zhe-Ming Lu.
\newblock Dea-net: Single image dehazing based on detail-enhanced convolution
  and content-guided attention.
\newblock \emph{IEEE Transactions on Image Processing}, 2024.

\bibitem[Dong et~al.(2020)Dong, Pan, Xiang, Hu, Zhang, Wang, and
  Yang]{dong2020multi}
Hang Dong, Jinshan Pan, Lei Xiang, Zhe Hu, Xinyi Zhang, Fei Wang, and
  Ming-Hsuan Yang.
\newblock Multi-scale boosted dehazing network with dense feature fusion.
\newblock In \emph{Proceedings of the IEEE/CVF conference on computer vision
  and pattern recognition}, pages 2157--2167, 2020.

\bibitem[Dong et~al.(2025)Dong, Vasa, Zhu, Qiu, Chen, Su, Xiong, Yang, Chen,
  and Wang]{dong2024cunsb}
Xuanzhao Dong, Vamsi~Krishna Vasa, Wenhui Zhu, Peijie Qiu, Xiwen Chen, Yi Su,
  Yujian Xiong, Zhangsihao Yang, Yanxi Chen, and Yalin Wang.
\newblock Cunsb-rfie: Context-aware unpaired neural schrodinger bridge in
  retinal fundus image enhancement.
\newblock In \emph{Proceedings of the Winter Conference on Applications of
  Computer Vision (WACV)}, pages 4502--4511, 2025.

\bibitem[Engin et~al.(2018)Engin, Gen{\c{c}}, and Kemal~Ekenel]{engin2018cycle}
Deniz Engin, Anil Gen{\c{c}}, and Hazim Kemal~Ekenel.
\newblock Cycle-dehaze: Enhanced cyclegan for single image dehazing.
\newblock In \emph{Proceedings of the IEEE conference on computer vision and
  pattern recognition workshops}, pages 825--833, 2018.

\bibitem[Fang et~al.(2025)Fang, Fan, Zheng, Weng, Tai, and Li]{fang2024sgdn}
Wenxuan Fang, Junkai Fan, Yu Zheng, Jiangwei Weng, Ying Tai, and Jun Li.
\newblock Guided real image dehazing using ycbcr color space.
\newblock In \emph{Proceedings of the AAAI Conference on Artificial
  Intelligence}, pages 2906--2914, 2025.

\bibitem[Farhadi and Redmon(2018)]{farhadi2018yolov3}
Ali Farhadi and Joseph Redmon.
\newblock Yolov3: An incremental improvement.
\newblock In \emph{Computer vision and pattern recognition}, pages 1--6.
  Springer Berlin/Heidelberg, Germany, 2018.

\bibitem[Fattal(2014)]{fattal2014dehazing}
Raanan Fattal.
\newblock Dehazing using color-lines.
\newblock \emph{ACM transactions on graphics (TOG)}, 34\penalty0 (1):\penalty0
  1--14, 2014.

\bibitem[Feng et~al.(2024)Feng, Ma, Meng, Zhou, Liu, and Su]{feng2024kanet}
Yuxin Feng, Long Ma, Xiaozhe Meng, Fan Zhou, Risheng Liu, and Zhuo Su.
\newblock Advancing real-world image dehazing: perspective, modules, and
  training.
\newblock \emph{IEEE Transactions on Pattern Analysis and Machine
  Intelligence}, 2024.

\bibitem[Goodfellow et~al.(2020)Goodfellow, Pouget-Abadie, Mirza, Xu,
  Warde-Farley, Ozair, Courville, and Bengio]{goodfellow2020gan}
Ian Goodfellow, Jean Pouget-Abadie, Mehdi Mirza, Bing Xu, David Warde-Farley,
  Sherjil Ozair, Aaron Courville, and Yoshua Bengio.
\newblock Generative adversarial networks.
\newblock \emph{Communications of the ACM}, 63\penalty0 (11):\penalty0
  139--144, 2020.

\bibitem[Guo et~al.(2024)Guo, Gao, Lu, Zhu, Liu, and He]{guo2024onerestore}
Yu Guo, Yuan Gao, Yuxu Lu, Huilin Zhu, Ryan~Wen Liu, and Shengfeng He.
\newblock Onerestore: A universal restoration framework for composite
  degradation.
\newblock In \emph{European Conference on Computer Vision}, pages 255--272.
  Springer, 2024.

\bibitem[He et~al.(2010)He, Sun, and Tang]{he2010single}
Kaiming He, Jian Sun, and Xiaoou Tang.
\newblock Single image haze removal using dark channel prior.
\newblock \emph{IEEE transactions on pattern analysis and machine
  intelligence}, 33\penalty0 (12):\penalty0 2341--2353, 2010.

\bibitem[Heusel et~al.(2017)Heusel, Ramsauer, Unterthiner, Nessler, and
  Hochreiter]{heusel2017gans}
Martin Heusel, Hubert Ramsauer, Thomas Unterthiner, Bernhard Nessler, and Sepp
  Hochreiter.
\newblock Gans trained by a two time-scale update rule converge to a local nash
  equilibrium.
\newblock \emph{Advances in neural information processing systems}, 30, 2017.

\bibitem[Ho et~al.(2020)Ho, Jain, and Abbeel]{ho2020denoising}
Jonathan Ho, Ajay Jain, and Pieter Abbeel.
\newblock Denoising diffusion probabilistic models.
\newblock \emph{Advances in neural information processing systems},
  33:\penalty0 6840--6851, 2020.

\bibitem[Ke et~al.(2021)Ke, Wang, Wang, Milanfar, and Yang]{ke2021musiq}
Junjie Ke, Qifei Wang, Yilin Wang, Peyman Milanfar, and Feng Yang.
\newblock Musiq: Multi-scale image quality transformer.
\newblock In \emph{Proceedings of the IEEE/CVF international conference on
  computer vision}, pages 5148--5157, 2021.

\bibitem[Kim et~al.(2024)Kim, Kwon, Kim, and Ye]{kim2023unpaired}
Beomsu Kim, Gihyun Kwon, Kwanyoung Kim, and Jong~Chul Ye.
\newblock Unpaired image-to-image translation via neural schrödinger bridge.
\newblock In \emph{The Twelfth International Conference on Learning
  Representations}, 2024.

\bibitem[Kumari et~al.(2022)Kumari, Zhang, Shechtman, and
  Zhu]{kumari2022ensembling}
Nupur Kumari, Richard Zhang, Eli Shechtman, and Jun-Yan Zhu.
\newblock Ensembling off-the-shelf models for gan training.
\newblock In \emph{Proceedings of the IEEE/CVF conference on computer vision
  and pattern recognition}, pages 10651--10662, 2022.

\bibitem[Lan et~al.(2025)Lan, Cui, Liu, Peng, Wang, Luo, and
  Liu]{lan2025exploiting}
Yunwei Lan, Zhigao Cui, Chang Liu, Jialun Peng, Nian Wang, Xin Luo, and Dong
  Liu.
\newblock Exploiting diffusion prior for real-world image dehazing with
  unpaired training.
\newblock In \emph{Proceedings of the AAAI Conference on Artificial
  Intelligence}, pages 4455--4463, 2025.

\bibitem[Li et~al.(2018)Li, Ren, Fu, Tao, Feng, Zeng, and
  Wang]{li2018benchmarking}
Boyi Li, Wenqi Ren, Dengpan Fu, Dacheng Tao, Dan Feng, Wenjun Zeng, and
  Zhangyang Wang.
\newblock Benchmarking single-image dehazing and beyond.
\newblock \emph{IEEE Transactions on Image Processing}, 28\penalty0
  (1):\penalty0 492--505, 2018.

\bibitem[Li et~al.(2021)Li, Gou, Gu, Liu, Zhou, and Peng]{li2021yoly}
Boyun Li, Yuanbiao Gou, Shuhang Gu, Jerry~Zitao Liu, Joey~Tianyi Zhou, and Xi
  Peng.
\newblock You only look yourself: Unsupervised and untrained single image
  dehazing neural network.
\newblock \emph{International Journal of Computer Vision}, 129:\penalty0
  1754--1767, 2021.

\bibitem[Li et~al.(2024)Li, Liao, Tang, Zhang, Li, Bian, Sheng, Feng, Li, Gao,
  et~al.]{li2024ustc}
Zhuoyuan Li, Junqi Liao, Chuanbo Tang, Haotian Zhang, Yuqi Li, Yifan Bian,
  Xihua Sheng, Xinmin Feng, Yao Li, Changsheng Gao, et~al.
\newblock Ustc-td: A test dataset and benchmark for image and video coding in
  2020s.
\newblock \emph{arXiv preprint arXiv:2409.08481}, 2024.

\bibitem[Liang et~al.(2023)Liang, Li, Zhou, Feng, and Loy]{liang2023iterative}
Zhexin Liang, Chongyi Li, Shangchen Zhou, Ruicheng Feng, and Chen~Change Loy.
\newblock Iterative prompt learning for unsupervised backlit image enhancement.
\newblock In \emph{Proceedings of the IEEE/CVF International Conference on
  Computer Vision}, pages 8094--8103, 2023.

\bibitem[Liu et~al.(2023)Liu, Vahdat, Huang, Theodorou, Nie, and
  Anandkumar]{liu2023i2sb}
Guan-Horng Liu, Arash Vahdat, De-An Huang, Evangelos~A. Theodorou, Weili Nie,
  and Anima Anandkumar.
\newblock I$^{2}$sb: Image-to-image schrödinger bridge.
\newblock In \emph{Proceedings of the 40th International Conference on Machine
  Learning}. JMLR.org, 2023.

\bibitem[Liu et~al.(2024)Liu, Ke, Liu, Zhao, and Lau]{liu2024diff}
Yuhao Liu, Zhanghan Ke, Fang Liu, Nanxuan Zhao, and Rynson~WH Lau.
\newblock Diff-plugin: Revitalizing details for diffusion-based low-level
  tasks.
\newblock In \emph{Proceedings of the IEEE/CVF Conference on Computer Vision
  and Pattern Recognition}, pages 4197--4208, 2024.

\bibitem[Mittal et~al.(2012)Mittal, Soundararajan, and Bovik]{mittal2012making}
Anish Mittal, Rajiv Soundararajan, and Alan~C Bovik.
\newblock Making a “completely blind” image quality analyzer.
\newblock \emph{IEEE Signal processing letters}, 20\penalty0 (3):\penalty0
  209--212, 2012.

\bibitem[Morawski et~al.(2024)Morawski, He, Dangi, and Hsu]{morawski2024}
Igor Morawski, Kai He, Shusil Dangi, and Winston~H Hsu.
\newblock Unsupervised image prior via prompt learning and clip semantic
  guidance for low-light image enhancement.
\newblock In \emph{Proceedings of the IEEE/CVF Conference on Computer Vision
  and Pattern Recognition}, pages 5971--5981, 2024.

\bibitem[Nayar and Narasimhan(1999)]{nayar1999vision}
Shree~K Nayar and Srinivasa~G Narasimhan.
\newblock Vision in bad weather.
\newblock In \emph{Proceedings of the seventh IEEE international conference on
  computer vision}, pages 820--827. IEEE, 1999.

\bibitem[Park et~al.(2020)Park, Efros, Zhang, and Zhu]{park2020cut}
Taesung Park, Alexei~A Efros, Richard Zhang, and Jun-Yan Zhu.
\newblock Contrastive learning for unpaired image-to-image translation.
\newblock In \emph{Computer Vision--ECCV 2020: 16th European Conference,
  Glasgow, UK, August 23--28, 2020, Proceedings, Part IX 16}, pages 319--345.
  Springer, 2020.

\bibitem[Qiu et~al.(2023)Qiu, Zhang, Wang, Luo, Li, and Jin]{qiu2023mb}
Yuwei Qiu, Kaihao Zhang, Chenxi Wang, Wenhan Luo, Hongdong Li, and Zhi Jin.
\newblock Mb-taylorformer: Multi-branch efficient transformer expanded by
  taylor formula for image dehazing.
\newblock In \emph{Proceedings of the IEEE/CVF International Conference on
  Computer Vision}, pages 12802--12813, 2023.

\bibitem[Radford et~al.(2021)Radford, Kim, Hallacy, Ramesh, Goh, Agarwal,
  Sastry, Askell, Mishkin, Clark, Krueger, and Sutskever]{clip}
Alec Radford, Jong~Wook Kim, Chris Hallacy, Aditya Ramesh, Gabriel Goh,
  Sandhini Agarwal, Girish Sastry, Amanda Askell, Pamela Mishkin, Jack Clark,
  Gretchen Krueger, and Ilya Sutskever.
\newblock {Learning Transferable Visual Models From Natural Language
  Supervision}.
\newblock In \emph{ICML}, pages 8748--8763, 2021.

\bibitem[Rombach et~al.(2022)Rombach, Blattmann, Lorenz, Esser, and
  Ommer]{rombach2022high}
Robin Rombach, Andreas Blattmann, Dominik Lorenz, Patrick Esser, and Bj{\"o}rn
  Ommer.
\newblock High-resolution image synthesis with latent diffusion models.
\newblock In \emph{Proceedings of the IEEE/CVF conference on computer vision
  and pattern recognition}, pages 10684--10695, 2022.

\bibitem[Ronneberger et~al.(2015)Ronneberger, Fischer, and
  Brox]{ronneberger2015u}
Olaf Ronneberger, Philipp Fischer, and Thomas Brox.
\newblock U-net: Convolutional networks for biomedical image segmentation.
\newblock In \emph{Medical image computing and computer-assisted
  intervention--MICCAI 2015: 18th international conference, Munich, Germany,
  October 5-9, 2015, proceedings, part III 18}, pages 234--241. Springer, 2015.

\bibitem[Schr{\"o}dinger(1932)]{schrodinger1932theorie}
Erwin Schr{\"o}dinger.
\newblock Sur la th{\'e}orie relativiste de l'{\'e}lectron et
  l'interpr{\'e}tation de la m{\'e}canique quantique.
\newblock In \emph{Annales de l'institut Henri Poincar{\'e}}, pages 269--310,
  1932.

\bibitem[Shao et~al.(2020)Shao, Li, Ren, Gao, and Sang]{shao2020domain}
Yuanjie Shao, Lerenhan Li, Wenqi Ren, Changxin Gao, and Nong Sang.
\newblock Domain adaptation for image dehazing.
\newblock In \emph{Proceedings of the IEEE/CVF conference on computer vision
  and pattern recognition}, pages 2808--2817, 2020.

\bibitem[Song et~al.(2021)Song, Meng, and Ermon]{song2021denoising}
Jiaming Song, Chenlin Meng, and Stefano Ermon.
\newblock Denoising diffusion implicit models.
\newblock In \emph{International Conference on Learning Representations}, 2021.

\bibitem[Wang et~al.(2024)Wang, Yan, Wang, Xie, Yang, Zhang, Qin, and
  Wei]{wang2024ucl}
Yongzhen Wang, Xuefeng Yan, Fu~Lee Wang, Haoran Xie, Wenhan Yang, Xiao-Ping
  Zhang, Jing Qin, and Mingqiang Wei.
\newblock Ucl-dehaze: Towards real-world image dehazing via unsupervised
  contrastive learning.
\newblock \emph{IEEE Transactions on Image Processing}, 2024.

\bibitem[Wang et~al.(2004)Wang, Bovik, Sheikh, and Simoncelli]{wang2004image}
Zhou Wang, Alan~C Bovik, Hamid~R Sheikh, and Eero~P Simoncelli.
\newblock Image quality assessment: from error visibility to structural
  similarity.
\newblock \emph{IEEE transactions on image processing}, 13\penalty0
  (4):\penalty0 600--612, 2004.

\bibitem[Yang et~al.(2022{\natexlab{a}})Yang, Wu, Shi, Lao, Gong, Cao, Wang,
  and Yang]{yang2022maniqa}
Sidi Yang, Tianhe Wu, Shuwei Shi, Shanshan Lao, Yuan Gong, Mingdeng Cao, Jiahao
  Wang, and Yujiu Yang.
\newblock Maniqa: Multi-dimension attention network for no-reference image
  quality assessment.
\newblock In \emph{Proceedings of the IEEE/CVF conference on computer vision
  and pattern recognition}, pages 1191--1200, 2022{\natexlab{a}}.

\bibitem[Yang et~al.(2022{\natexlab{b}})Yang, Wang, Liu, Zhang, Guo, and
  Tao]{yang2022self}
Yang Yang, Chaoyue Wang, Risheng Liu, Lin Zhang, Xiaojie Guo, and Dacheng Tao.
\newblock Self-augmented unpaired image dehazing via density and depth
  decomposition.
\newblock In \emph{Proceedings of the IEEE/CVF conference on computer vision
  and pattern recognition}, pages 2037--2046, 2022{\natexlab{b}}.

\bibitem[Yang et~al.(2024)Yang, Wang, Guo, and Tao]{yang2024robust}
Yang Yang, Chaoyue Wang, Xiaojie Guo, and Dacheng Tao.
\newblock Robust unpaired image dehazing via density and depth decomposition.
\newblock \emph{International Journal of Computer Vision}, 132\penalty0
  (5):\penalty0 1557--1577, 2024.

\bibitem[Zhang et~al.(2014)Zhang, Shen, and Li]{zhang2014vsi}
Lin Zhang, Ying Shen, and Hongyu Li.
\newblock Vsi: A visual saliency-induced index for perceptual image quality
  assessment.
\newblock \emph{IEEE Transactions on Image processing}, 23\penalty0
  (10):\penalty0 4270--4281, 2014.

\bibitem[Zhang et~al.(2018)Zhang, Isola, Efros, Shechtman, and
  Wang]{zhang2018unreasonable}
Richard Zhang, Phillip Isola, Alexei~A Efros, Eli Shechtman, and Oliver Wang.
\newblock The unreasonable effectiveness of deep features as a perceptual
  metric.
\newblock In \emph{Proceedings of the IEEE conference on computer vision and
  pattern recognition}, pages 586--595, 2018.

\bibitem[Zhao et~al.(2021)Zhao, Zhang, Shen, and Zhou]{zhao2021refinednet}
Shiyu Zhao, Lin Zhang, Ying Shen, and Yicong Zhou.
\newblock Refinednet: A weakly supervised refinement framework for single image
  dehazing.
\newblock \emph{IEEE Transactions on Image Processing}, 30:\penalty0
  3391--3404, 2021.

\bibitem[Zheng et~al.(2023)Zheng, Zhan, He, Dong, and Du]{zheng2023curricular}
Yu Zheng, Jiahui Zhan, Shengfeng He, Junyu Dong, and Yong Du.
\newblock Curricular contrastive regularization for physics-aware single image
  dehazing.
\newblock In \emph{Proceedings of the IEEE/CVF conference on computer vision
  and pattern recognition}, pages 5785--5794, 2023.

\bibitem[Zhou et~al.(2019)Zhou, Zhao, Puig, Xiao, Fidler, Barriuso, and
  Torralba]{zhou2019semantic}
Bolei Zhou, Hang Zhao, Xavier Puig, Tete Xiao, Sanja Fidler, Adela Barriuso,
  and Antonio Torralba.
\newblock Semantic understanding of scenes through the ade20k dataset.
\newblock \emph{International Journal of Computer Vision}, 127:\penalty0
  302--321, 2019.

\bibitem[Zhu et~al.(2017)Zhu, Park, Isola, and Efros]{zhu2017unpaired}
Jun-Yan Zhu, Taesung Park, Phillip Isola, and Alexei~A Efros.
\newblock Unpaired image-to-image translation using cycle-consistent
  adversarial networks.
\newblock In \emph{Proceedings of the IEEE international conference on computer
  vision}, pages 2223--2232, 2017.

\bibitem[Zhu et~al.(2015)Zhu, Mai, and Shao]{zhu2015fast}
Qingsong Zhu, Jiaming Mai, and Ling Shao.
\newblock A fast single image haze removal algorithm using color attenuation
  prior.
\newblock \emph{IEEE transactions on image processing}, 24\penalty0
  (11):\penalty0 3522--3533, 2015.

\end{thebibliography}
}
\end{document}